\documentclass{article}

\usepackage{template/preprint,times}

\usepackage{amsmath,amsfonts,bm}

\def\eqref#1{equation~\ref{#1}}

\def\1{\bm{1}}

\DeclareMathAlphabet{\mathsfit}{\encodingdefault}{\sfdefault}{m}{sl}
\SetMathAlphabet{\mathsfit}{bold}{\encodingdefault}{\sfdefault}{bx}{n}

\usepackage{booktabs}
\usepackage{multirow}
\usepackage{microtype}
\usepackage{xspace}
\usepackage{graphicx}
\usepackage{float}
\usepackage{xcolor}
\usepackage{amsmath,amssymb,amsthm}
\usepackage{algorithm}
\usepackage{algorithmic}
\usepackage{placeins}
\usepackage{hyperref}
\usepackage{url}
\usepackage{fontawesome5}

\definecolor{preprintBlue}{HTML}{285E95}
\definecolor{preprintTeal}{HTML}{197D74}
\hypersetup{
  colorlinks=true,
  linkcolor=preprintBlue,
  citecolor=preprintTeal,
  urlcolor=preprintBlue,
  pdfborder={0 0 0},
  pdftitle={VPEvolve: A Self-Evolving Virtual Process Engineer for Computational Lithography},
  pdfauthor={Tianyi Li, Wenxuan Dong, Donger Luo, Nan Wang, Yanpeng Chen, Jiaqi Liu, Xinyun Zhang, Hao Geng}
}

\newcommand{\system}{\textsc{VPEvolve}\xspace}

\theoremstyle{definition}

\title{\system: A Self-Evolving Virtual Process Engineer for Computational Lithography}

\author{
\textbf{Tianyi Li}\textsuperscript{1,2}\thanks{Equal contribution.} \quad
\textbf{Wenxuan Dong}\textsuperscript{1,2}\footnotemark[1] \quad
\textbf{Donger Luo}\textsuperscript{1} \quad
\textbf{Nan Wang}\textsuperscript{1,2} \\[1pt]
\textbf{Yanpeng Chen}\textsuperscript{1} \quad
\textbf{Jiaqi Liu}\textsuperscript{3} \quad
\textbf{Xinyun Zhang}\textsuperscript{1} \quad
\textbf{Hao Geng}\textsuperscript{1}\thanks{Corresponding author.}
\\[2mm]
{\normalfont\small \textsuperscript{1}ShanghaiTech University \quad
\textsuperscript{2}ZeroShot Co., Ltd.} \\[1pt]
{\normalfont\small \textsuperscript{3}Shanghai Optoelectronics Science and Technology Innovation Center} \\[2pt]
{\normalfont\footnotesize \texttt{\{lity2024, dongwx2025, genghao\}@shanghaitech.edu.cn}}
}

\begin{document}
\maketitle

{\centering\small
\href{https://true-litianyi.github.io/VPEVOLVE/}{\textcolor{preprintBlue}{\faGlobe}\;\textbf{Website}}
\hspace{1.5em}
\href{https://github.com/True-Litianyi/VPEVOLVE_code}{\textcolor{preprintTeal}{\faGithub}\;\textbf{GitHub}}\par}
\vspace{4pt}

\begin{abstract}
Optical proximity correction (OPC) recipes grow as engineers add local rules to repair newly discovered lithography hotspots. Each correction can interact with existing rules, while lessons from commercial-tool trials remain scattered across code and logs. \system combines a Virtual Process Engineer (VPE) harness with a Skill Bank of measured engineering experience. The harness equips a frozen language model with process manuals, layout analysis, recipe editing, and commercial-tool evaluation. The actor proposes changes to the global parameters, local targeted rules, or diagnostic trials. After each evaluation, an LLM reflector and curator turn the measured response into evidence-linked judgments. The actor retrieves them before its next trial. Feasible improvements update the retained recipe; every measured trial informs the Skill Bank that guides the next edit. The model weights remain fixed. On a FreePDK45-derived benchmark with ten commercial-tool evaluations per case, \system reduces the mean per-case maximum edge placement error from 18.294 to 5.361 nm on Poly and from 22.052 to 15.692 nm on Metal1. Every final recipe satisfies the predefined quality constraints and improves the maximum error by at least 0.1 nm.
\end{abstract}

\begin{figure}[H]
\centering
\includegraphics[width=\textwidth]{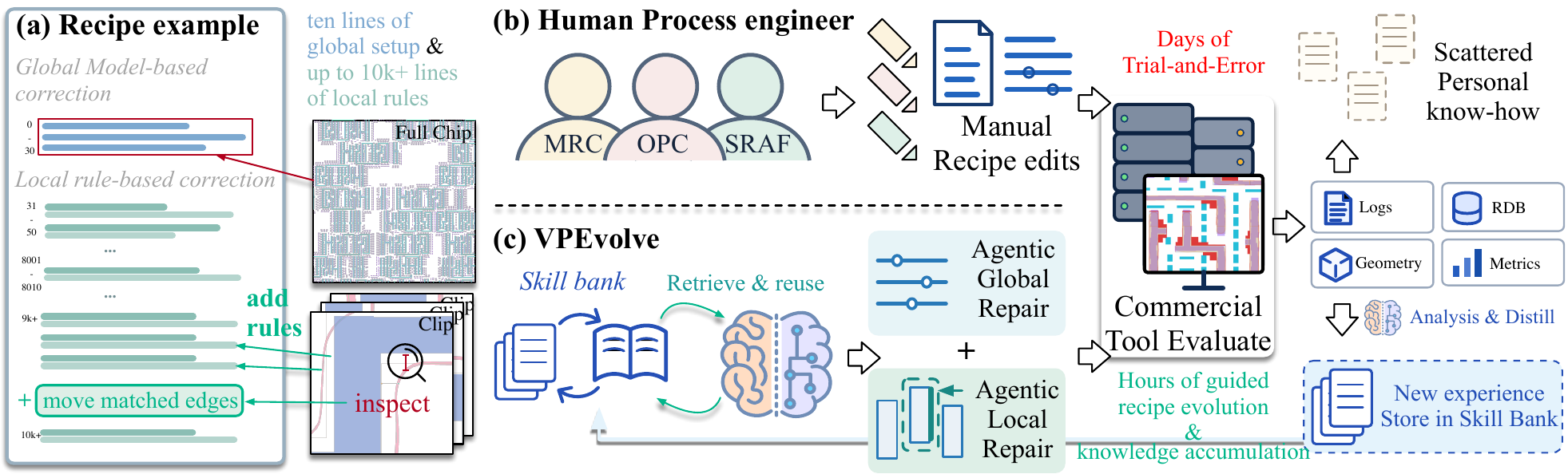}
\caption{\textbf{From recipe development to VPEvolve.}
(a) OPC recipes combine global settings with local rules for specific patterns.
(b) Engineers diagnose printing errors and revise recipes, accumulating
experience across edits and trial records. (c) VPEvolve connects recipe
optimization with a Skill Bank that turns measured outcomes into guidance
for subsequent experiments.}
\label{fig:vpevolve-paradigm}
\end{figure}

\section{Introduction}
\label{sec:introduction}

In semiconductor manufacturing, optical proximity effects distort the shapes
printed on a wafer \citep{mack2007lithography}. Process engineers therefore
develop a \textbf{recipe} for each process node and target layer to control how
an optical proximity correction (OPC) tool corrects the mask. A recipe is more
than a set of model parameters. Model-based correction may require only a few
dozen lines of settings, while thousands or tens of thousands of lines encode
local rules for specific layout patterns, as illustrated in
Figure~\ref{fig:vpevolve-paradigm}(a). When simulation or test-chip
measurements reveal large edge placement error (EPE) or mask rule violation, engineers inspect the affected geometry,
write rules to identify similar patterns, and specify how to correct them.
The tool applies those corrections wherever the patterns recur. As rules
accumulate, reading and changing the recipe becomes time-consuming: a fix for
one hotspot can interact with existing rules or create another violation.
Engineers responsible for OPC, sub-resolution assist features (SRAF) and mask rule checking (MRC)
must coordinate edits and repeatedly test the results, as illustrated in
Figure~\ref{fig:vpevolve-paradigm}(b). However, the link between what changed, why,
and what happened remains scattered across engineers, code, and logs, making experience hard to reuse.

Traditional parameter-tuning methods cover only part of recipe development. Bayesian optimization can tune model-based settings chosen in advance \citep{snoek2012practical}. But process engineers often add local rules after seeing new hotspots, introducing pattern matching conditions and correction values that were not in the original search space. Large language models (LLMs) can propose such edits, but recipe work also requires process knowledge and access to the layout geometry behind a measured error. The lesson of a successful edit can become misleading as the recipe changes. A local rule may fix its target while another hotspot becomes the worst, and later global edits may change the effect of the rule. If the agent only remembers that the rule worked, it can keep refining that correction after the limiting error has moved elsewhere. This motivates a \textbf{Virtual Process Engineer (VPE) harness} that gives the agent process
knowledge, layout analysis, recipe editing, and commercial-tool feedback.
\system combines this harness with an LLM-curated Skill Bank that turns
trial outcomes into reusable engineering judgments (Figure~\ref{fig:vpevolve-paradigm}(c)).

The VPE harness supplies process manuals for recipe commands and editing
procedures. The LLM actor reads the current recipe, layout diagnostics, and
previous outcomes, then chooses a global edit, a pattern-specific local
correction, or a diagnostic trial. Layout tools identify the relevant edges
and compile the edit; the commercial tool measures the printing and
manufacturing response \citep{siemenscalibre}. After each trial, an LLM
reflector interprets local and global measurements. An LLM curator revises
the Skill Bank with the conditions, supporting trials, counterexamples, and
possible next experiments for each judgment. The actor retrieves relevant
judgments alongside the latest feedback before deciding what to try next.
Measured improvements update the retained recipe; every measured trial
supplies evidence for the next edit.

Recipe development is evaluated on a FreePDK45-derived benchmark with ten
commercial-tool evaluations per case. \system reduces the mean per-case maximum EPE
from 18.294 to 5.361 nm on Poly and from 22.052 to 15.692 nm on Metal1; every endpoint
meets the predefined quality constraints and improves by at least 0.1 nm.
Mechanism ablations examine how interpreting spatial responses, revising
experience, and guiding subsequent trials contribute to these improvements.
To distinguish better recipes from more useful experience, a paired
continuation study holds the recipe and observed history fixed while varying
how experience is curated. Further analyses examine local actions and improvement over longer search horizons.
Our contributions:
\begin{itemize}
    \item \textbf{Recipe development as a learning problem.} The task is formulated as sequential search over global settings and growing local rules, with trial evidence retained for later decisions.
    \item \textbf{A Virtual Process Engineer harness.} Process manuals, layout analysis, recipe editing, and commercial-tool evaluation equip a frozen LLM to run global, local, and diagnostic trials.
    \item \textbf{Self-evolving engineering experience.} LLM reflection and curation distill each trial into scoped, evidence-linked skills that guide subsequent experiments.
    \item \textbf{Evaluation and an open testbed.} Commercial-tool experiments show lower maximum EPE than all seven baselines on both layers. Ablations, language-model comparisons, and longer runs test what drives the gain and when it persists. The FreePDK45-derived layouts and lithography simulator are released for open simulation studies.
\end{itemize}

\section{Preliminaries}
\label{sec:preliminaries}

\begin{figure}[t]
\centering
\includegraphics[width=\textwidth]{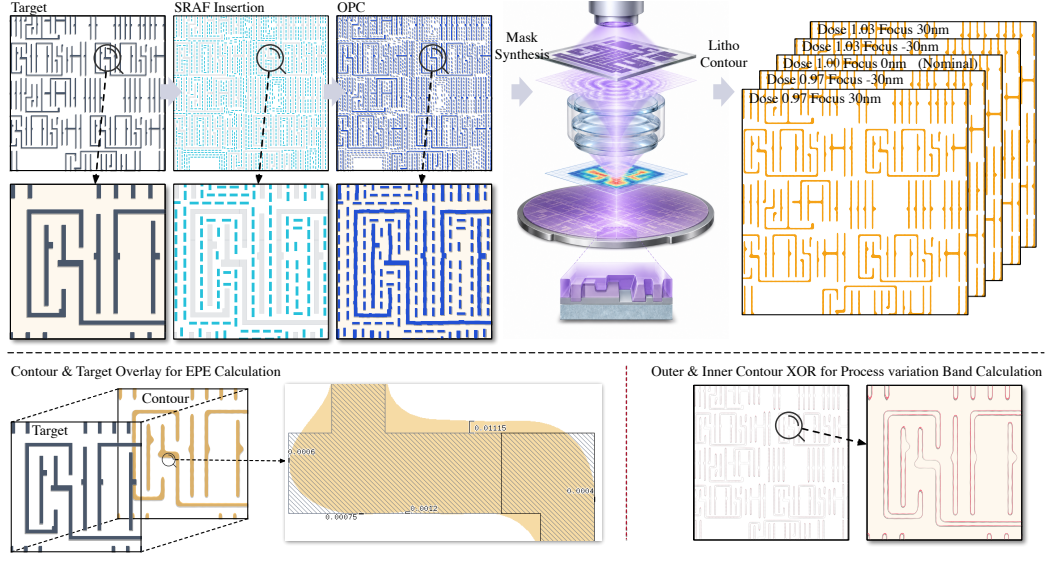}
\caption{\textbf{Computational-lithography evaluation.}
A recipe guides SRAF insertion and OPC to produce a mask, whose printed
contours are simulated across focus and exposure conditions. EPE compares
printed contours with target edges; PVB measures their variation across
process conditions. These measurements guide subsequent recipe edits.}
\label{fig:computational-lithography-example}
\end{figure}

An edit to a recipe is judged by what it prints. Given a layout and a recipe,
the lithography tool generates a mask and predicts the wafer contours at
different focus and exposure-dose settings, known as process corners
\citep{mack2007lithography,yang2023physicslitho}
(Figure~\ref{fig:computational-lithography-example}). The recipe controls
optical proximity correction (OPC), which reshapes mask features to
compensate for proximity effects, and places sub-resolution assist features
(SRAFs) to stabilize imaging without printing on the wafer.

Recipe quality is measured from these contours. At a target-edge sample
$p_i$, let $n_i$ be the unit normal and $q_{i,c}$ the corresponding contour
point at process corner $c\in\mathcal C$. With $P_c$ the printed region and
$\mu$ denoting area, define edge-placement error (EPE) and the process
variation band (PVB) by
\begin{equation}
\mathrm{EPE}_{i,c}=\langle q_{i,c}-p_i,n_i\rangle,
\qquad
\mathrm{PVB}=\frac{1}{L}\mu\!\left[
\left(\bigcup_{c\in\mathcal C}P_c\right)
\setminus\left(\bigcap_{c\in\mathcal C}P_c\right)
\right],
\label{eq:epe-pvb}
\end{equation}
The sign of EPE indicates the direction of contour displacement. Reported
Max and Mean EPE are the maximum and mean absolute errors at the fixed
evaluation gauges under nominal focus and dose. Top-10 EPE averages the
absolute errors of the ten highest-error, spatially separated hotspots
selected by the evaluator. PVB measures variation across process corners,
normalized by the target-edge length $L$ in the scoring region, and has
units of $\mu$m. Mask rule checks (MRC) assess mask manufacturability, while
SRAF printability checks detect assist features that print unintentionally.
These measurements determine whether a recipe can be retained and identify
residual errors.

Repeated evaluation makes this an optimization problem, but the choices do
not stay the same throughout recipe development. Bayesian optimization
searches an expensive parameter space specified in advance
\citep{snoek2012practical,gardner2014constrainedbo}. Learning-based
lithography methods have produced mask corrections and SRAFs
\citep{yang2018ganopc,chen2020damo,yang2024ililt,li2025llmsraf}. Chen et
al.\ choose EPE measurement positions and edge fragmentation for
model-based OPC with reinforcement learning, then distill the results into
pattern-conditioned rules using multimodal feature labels and a decision
tree \citep{chen2025opcengineer}. A recipe also grows through new local
rules chosen in response to measured hotspots. Each new rule expands the
available edits and can change how previous trial outcomes inform subsequent
decisions.

Agents provide a way to work through this changing sequence of tool calls
and edits. Methods for tool use interleave reasoning with execution and
learn when to call an API \citep{schick2023toolformer,yao2023react}; agents
for science and EDA coordinate experiments and engineering tools
\citep{liu2024agenthpo,huang2024mlagentbench,zhang2026scinav,wu2024chateda}.
ChipNeMo adapts language models for chip-design assistance and script
generation \citep{liu2023chipnemo}.

After a trial, the measured result can inform a later decision even if the
candidate recipe is rejected. Reflection, memory, and workflows provide
ways to carry such information across attempts
\citep{shinn2023reflexion,zhao2024expel,wang2025awm}. Prior agents have
stored reusable behavior as executable skills \citep{wang2023voyager} or
contextual strategies and playbooks
\citep{ouyang2026reasoningbank,zhang2026ace}; the Darwin G\"odel Machine
also modifies agent code \citep{zhang2026dgm}. In \system, the recipe and
the engineering judgments used to edit it both evolve from measured
commercial-tool feedback, while the agent harness stays fixed.

\section{Method}
\label{sec:method}
\suppressfloats[t]

\begin{figure}[t]
\centering
\includegraphics[width=\textwidth]{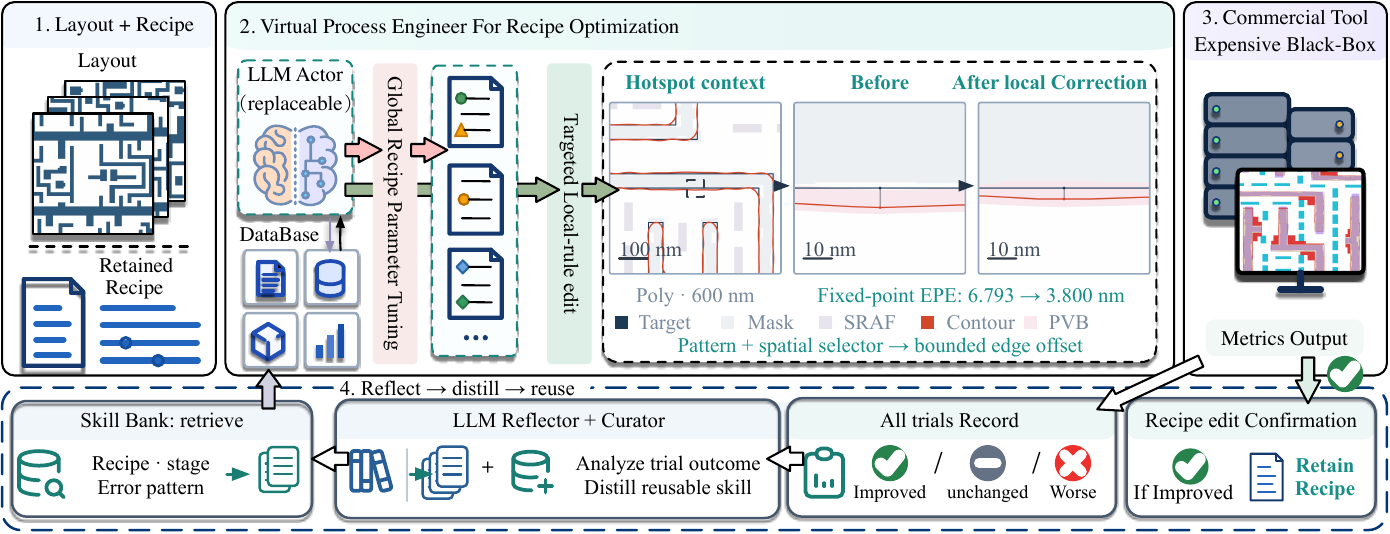}
\caption{\textbf{Recipe and experience evolution in VPEvolve.}
The actor uses the current recipe, spatial feedback, and retrieved experience
to propose global changes or local corrections for commercial-tool evaluation.
Feasible improvements update the retained recipe. The reflector interprets the
measured response, and the curator revises the Skill Bank to guide subsequent
trials; model weights remain fixed throughout both updates. The Poly example
shows a local correction reducing EPE at the fixed measurement point from
6.793 to 3.800 nm.}
\label{fig:vpevolve-harness}
\end{figure}

\subsection{Recipe Development as a Sequential Decision Problem}
\label{sec:problem}

Each recipe trial reveals how a candidate prints an IC layout and informs the
next modification. This makes recipe development a sequential decision problem.
Given a layout $x$, an initial recipe $r_0$, and a budget of $B$ optimization
trials, the objective is to reduce the maximum edge-placement error while
maintaining acceptable average EPE, process variation, and mask
manufacturability. The process model and evaluation settings remain fixed
during the search.

\paragraph{Actions and feedback.}
At step $t$, the agent uses the retained recipe $r_t$ and its simulation
feedback $e_t$ to choose a global edit, a pattern-specific local correction, or a
diagnostic trial. Global edits adjust recipe parameters that affect the layout
as a whole; local edits introduce correction rules for selected geometric
patterns where printing errors remain. A diagnostic trial evaluates a
controlled recipe change to test a hypothesis about the measured response.
Evaluating a
candidate recipe returns both overall quality metrics and the locations of
residual errors. A candidate replaces the current recipe when it satisfies
the quality constraints and improves the optimization objective. Otherwise,
the current recipe is retained, and the observed outcome informs the next
decision.

\paragraph{Learning from recipe development.}
Each evaluation records what changed, under which conditions, and how the
printed result responded. A trial record $\tau_t$
captures the outcome, and a versioned bank $\mathcal E_t$ maintains
engineering judgments. Before proposing a modification, the
actor retrieves relevant judgments. After evaluation, an LLM reflector
interprets the measured response, and an LLM curator proposes a revision to
the bank. The resulting loop is
\begin{equation}
\begin{aligned}
V_t &= \operatorname{Retrieve}(\mathcal E_t,x,r_t,e_t),
& a_t &\sim \pi^{\mathrm{act}}_{\theta,H}(\cdot\mid x,r_t,e_t,V_t,\mathcal T_t),\\
\widetilde r_t &= \operatorname{Apply}(r_t,a_t),
& \widetilde e_t &= F(x,\widetilde r_t),\\
(r_{t+1},e_{t+1}) &= \operatorname{Select}(r_t,e_t,\widetilde r_t,\widetilde e_t),
& h_t &= \operatorname{Reflect}_{\pi_\theta}(\tau_t),\\
\Delta_t &= \operatorname{Curate}_{\pi_\theta}(\mathcal E_t,h_t,\tau_t),
& \mathcal E_{t+1} &= \operatorname{CheckApply}(\mathcal E_t,\Delta_t,\tau_t).
\end{aligned}
\label{eq:harness-evolution}
\end{equation}
Here, $H$ denotes the fixed agent harness, $V_t$ retrieved experience,
$a_t$ the actor's proposed trial, and $F$ the fixed lithography evaluator.
Selection returns the candidate
recipe and its feedback when it is a feasible improvement, and retains
$(r_t,e_t)$ otherwise. The trial record
$\tau_t=(r_t,a_t,\widetilde r_t,e_t,\widetilde e_t)$ is retained even when the
candidate recipe is rejected and appended to the trial history
$\mathcal T_{t+1}=\mathcal T_t\cup\{\tau_t\}$, with $\mathcal T_0=\varnothing$.
The reflection $h_t$ distinguishes measured responses
from hypotheses; the curator's patch $\Delta_t$ may add, update, split, or
merge judgments and suggest a subsequent experiment. The harness checks
evidence references and revision structure before applying that patch. Recipe
and experience can change independently: even a rejected candidate can reveal
where a rule works or fails. The model weights, process model, and harness
remain fixed.

\subsection{Virtual Process Engineer Harness}

\system retains the best feasible executable recipe and a versioned bank of
engineering judgments. Its VPE harness connects an LLM actor to process
guidance, layout analysis, recipe editing, and commercial-tool evaluation.
An LLM reflector analyzes each measured trial, and a curator revises the
actor's experience. All three roles share one frozen model
(Figure~\ref{fig:vpevolve-harness}).

The actor reads the current recipe, spatial and aggregate feedback, previous
trial records, process manuals, and retrieved experience. It chooses
whether to adjust global settings, apply a local displacement rule, or perform a
diagnostic experiment. The process manuals supply recipe commands and
engineering strategies. The actor revises its strategy as evidence arrives.
Each trial uses one commercial-tool evaluation from the shared budget.

\begin{samepage}
For a local action, layout tools associate a measured EPE hotspot with its
design edge and nearby geometry. Native pattern matching identifies eligible
mask-edge fragments, and a spatial selector binds the proposed displacement
to the diagnosed occurrence. The EPE sign and edge normal provide a starting
direction; the actor tests and revises that hypothesis against the measured
response. The LLM chooses the experiment; a deterministic compiler binds
every edit to its exact parent recipe and checks allowed commands, geometry,
and displacement limits before execution.
\par
\end{samepage}

The commercial tool returns EPE at the original hotspot and newly worst locations,
global Max/Top-10/Mean EPE, PVB, MRC, and SRAF printability checks. A candidate
becomes the next recipe only if it satisfies the fixed constraints and improves
the prescribed lexicographic quality ordering. Every trial is recorded
together with its measurements or execution outcome and the evaluated recipe.

\subsection{Self-Evolving Skill Bank}

The experience mechanism adapts ACE-style reflection and curation
\citep{zhang2026ace} to spatial process feedback. It adds three connected
capabilities. \textbf{Domain Response Reflection} compares the intended
effect of a trial with the measured response at its original hotspot, the
new worst hotspot, and the full scoring region. The LLM distinguishes no response from a repaired target followed by
a new global bottleneck, and records possible side effects as hypotheses
rather than established facts.

\textbf{Scoped Experience Revision} lets the LLM curator add, update,
split, or merge an engineering judgment. A judgment links an intervention and
its applicability conditions to observed effects, supporting trials,
counterexamples, and unresolved questions. The curator explains changes to
evidence roles. Programmatic checks verify that cited trials exist, that an
intervention occurred in them, and that the versioned patch respects its
declared structure. The LLM assesses applicability and causal interpretation
from the measured record. For example, when a rule repairs its target but
another hotspot becomes limiting, the curator can preserve evidence of
local improvement while qualifying its effect on maximum EPE. A reversed
response under changed recipe settings can provide counterevidence and
narrow the judgment's applicability.

\textbf{Experiment Guidance} allows the curator to attach
a strategy update and up to two next proposed experiments. Each proposal
states an action, its expected local and global effects, and an observation
that would falsify it. These suggestions enter the actor's next context; the
actor chooses the next action. Retrieval favors judgments
relevant to the current recipe, the layer, and the measured response. Original
trial records remain available alongside the curated bank. The actor uses
these records and distilled findings in subsequent decisions without changing
model weights or harness code. Appendix~\ref{app:implementation}
details the implementation.

\section{Experiments}
\label{sec:experiments}

\subsection{Experimental Setup}

\begin{table}[t]
\caption{\textbf{VPEvolve versus seven baselines on Poly and Metal1.}
Layer-wise EPE and PVB are case means; overall success and call counts cover
the benchmark. EPE is in nm and PVB in
$10^{-3}\,\mu$m.
}
\label{tab:current-online}
\centering\footnotesize
\setlength{\tabcolsep}{2pt}\renewcommand{\arraystretch}{1.08}
\begin{tabular*}{\textwidth}{@{\extracolsep{\fill}}lrrrrrrrrr@{}}
\toprule
& \multicolumn{3}{c}{Poly} & \multicolumn{3}{c}{Metal1} & \multicolumn{3}{c}{Overall}\\
\cmidrule(lr){2-4}\cmidrule(lr){5-7}\cmidrule(l){8-10}
Method & Max$\downarrow$ & Avg.$\downarrow$ & PVB$\downarrow$ & Max$\downarrow$ & Avg.$\downarrow$ & PVB$\downarrow$ & Max$\downarrow$ & SR$\uparrow$ & OPC / LLM\\
\midrule
Initial recipe & 18.294 & 0.854 & 7.281 & 22.052 & 0.805 & 5.015 & 20.173 & \textemdash & \textemdash\\
\addlinespace[2pt]
Raw Actor & 17.271 & 0.835 & 7.239 & 19.609 & 0.758 & 4.969 & 18.440 & 80\% & 220 / 455\\
ReAct & 16.501 & 0.817 & 7.120 & 17.775 & 0.729 & 4.885 & 17.138 & 100\% & 220 / 524\\
Textual Memory & 17.296 & 0.781 & 6.977 & 18.966 & 0.742 & 4.878 & 18.131 & 85\% & 220 / 1030\\
ReasoningBank & 17.523 & 0.833 & 7.151 & 20.538 & 0.756 & 4.924 & 19.031 & 75\% & 220 / 988\\
ACE & 16.697 & 0.783 & 6.905 & 19.920 & 0.758 & 4.928 & 18.308 & 95\% & 220 / 1833\\
Static Harness & 17.583 & 0.848 & 7.257 & 19.875 & 0.780 & 5.001 & 18.729 & 75\% & 220 / 502\\
Bayesian opt. & 17.909 & 0.827 & 7.300 & 21.773 & 0.778 & 5.010 & 19.841 & 30\% & 220 / 0\\
\midrule
VPEvolve & \textbf{5.361} & 0.645 & 7.398 & \textbf{15.692} & 0.690 & 5.021 & \textbf{10.527} & 100\% & 220 / 1295\\
\bottomrule
\end{tabular*}
\par\vspace{2pt}\begin{minipage}{\textwidth}\footnotesize
SR: feasible max-EPE reduction $\geq0.1$ nm; all retained endpoints have MRC = 0.
OPC counts exclude R0 and include endpoint repeats. Bold marks VPEvolve;
Appendix~\ref{app:runtime} details evaluation and model-call accounting.
\end{minipage}
\end{table}

\paragraph{Benchmark and evaluation.}
Each commercial-tool evaluation runs recipe correction and lithography
simulation under the fixed process model.
The benchmark has 20 windows cropped from FreePDK45-generated full-chip
layouts \citep{freepdk45}: ten Poly and ten Metal1, balanced across five
density strata. Each has a $10\times10\,\mu$m scoring region and a $2\,\mu$m
context halo. Methods share R0, the process model, and measurement points.
VPEvolve and its mechanism ablations start with an empty Skill Bank and
shared engineering guidance. Each case has one seed, a ten-trial optimization budget,
and a final recipe repeat. The objective minimizes maximum EPE while mean EPE and PVB remain
within 102\% of R0, with zero MRC and SRAF-printing violations. Ties are
broken by top-ten mean EPE, mean EPE, then PVB. We report maximum and mean
EPE, PVB, and success by layer; success requires completion and a feasible
maximum-EPE reduction of at least 0.1 nm. Appendices~\ref{app:dataset},
\ref{app:platform}, and~\ref{app:runtime} give metric and cost details.

\paragraph{Baselines.}
We compare with six agent baselines and Bayesian optimization.
Raw Actor, ReAct \citep{yao2023react}, and Static Harness
\citep{pi2026agent} test action selection and fixed guidance.
Textual Memory \citep{shinn2023reflexion,zhao2024expel},
ReasoningBank \citep{ouyang2026reasoningbank}, and ACE
\citep{zhang2026ace} use reflection, retrieval, or context revision.
Bayesian optimization \citep{snoek2012practical,gardner2014constrainedbo}
searches model-based parameters. Agent baselines share recipe-editing
actions and trial budgets but vary in context and memory. These baselines cover numerical search,
general-purpose agents, and memory mechanisms (Appendix~\ref{app:baselines}).

\paragraph{Implementation.}
VPEvolve uses the Pi agent harness \citep{pi2026agent} and extends ACE-style
reflection and curation \citep{zhang2026ace} with domain response reflection,
scoped experience revision, and experiment guidance. Unless stated otherwise,
all three roles use local Qwen3.6-27B \citep{qwenteam2026qwen36} on four
NVIDIA H100 GPUs, with a 131,072-token context, 4,096 output tokens, and
temperature zero; model weights remain fixed.
Cross-model experiments use each evaluated model's API for all three roles;
long-horizon experiments use the DeepSeek-V4.1-Flash API.
Appendix~\ref{app:implementation} gives the inference settings.

\subsection{Main Results}
\label{sec:current-online}

Table~\ref{tab:current-online} tests whether VPEvolve improves recipe quality
under a limited evaluation budget. It reduces mean window-maximum EPE from
18.294 to 5.361 nm on Poly and from 22.052 to 15.692 nm on Metal1, reductions
of 70.7\% and 28.8\%. All cases achieve a feasible improvement. ReAct is the
strongest agent baseline on maximum EPE, at 16.501 and 17.775 nm; VPEvolve
improves on it by 67.5\% and 11.7\%. Mean EPE also falls below every baseline,
to 0.645 and 0.690 nm. VPEvolve thus improves worst-case and average accuracy within the fixed
quality constraints.

\begin{figure}[t]
\centering
\includegraphics[width=\textwidth]{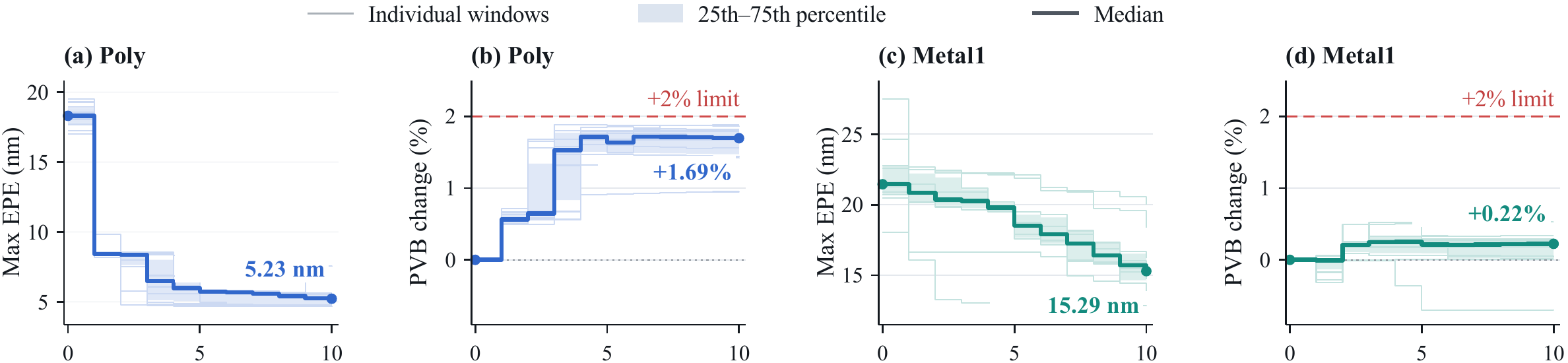}
\caption{\textbf{Recipe evolution over ten optimization trials.}
Maximum EPE and PVB change from R0 for retained recipes. Thin lines are cases;
thick lines and bands show medians and interquartile ranges. Endpoint labels
are medians; Table~\ref{tab:current-online} reports means. The dashed line marks
the +2\% PVB limit. Trial zero is R0; endpoint repeats are excluded.}
\label{fig:main-trajectories}
\end{figure}

Figure~\ref{fig:main-trajectories} shows how these gains develop. On Poly,
median maximum EPE drops to 8.42 nm after one trial and reaches 5.23 nm after
ten. Metal1 improves later, from 18.52 nm at trial five to 15.29 nm at trial
ten. Refinement continues after the initial edits while PVB stays within
the +2\% limit.

\begin{table}[t]
\caption{\textbf{Mechanism ablation of experience evolution.}
Metrics average the best feasible retained recipe for each case in each layer.
EPE is in nm and PVB in $10^{-3}\,\mu$m. SR is the share of assigned runs
completing with a feasible maximum-EPE reduction of at least 0.1 nm.}
\label{tab:experience-ablation}
\centering\footnotesize
\setlength{\tabcolsep}{3pt}\renewcommand{\arraystretch}{1.08}
\begin{tabular*}{\textwidth}{@{\extracolsep{\fill}}lrrrrrrrr@{}}
\toprule
& \multicolumn{4}{c}{Poly} & \multicolumn{4}{c}{Metal1}\\
\cmidrule(lr){2-5}\cmidrule(l){6-9}
Condition & Max$\downarrow$ & Mean$\downarrow$ & PVB$\downarrow$ & SR$\uparrow$
& Max$\downarrow$ & Mean$\downarrow$ & PVB$\downarrow$ & SR$\uparrow$\\
\midrule
\textbf{Full VPEvolve} & 5.361 & 0.645 & 7.398 & 100\% & 15.692 & 0.690 & 5.021 & 100\%\\
Frozen Experience & 8.082 & 0.640 & 7.319 & 90\% & 16.121 & 0.717 & 5.027 & 100\%\\
w/o Domain Response Reflection & 6.213 & 0.653 & 7.374 & 90\% & 16.280 & 0.742 & 5.023 & 100\%\\
w/o Scoped Experience Revision & 5.475 & 0.632 & 7.411 & 100\% & 16.334 & 0.717 & 5.025 & 90\%\\
w/o Experiment Guidance & 5.870 & 0.624 & 7.394 & 100\% & 16.154 & 0.725 & 5.025 & 90\%\\
\bottomrule
\end{tabular*}
\end{table}

\subsection{Mechanism Ablation}
\label{sec:experience-ablation}

Table~\ref{tab:experience-ablation} examines experience accumulation and its
three components. Full VPEvolve and four variants share the model, seeds,
R0, empty initial bank, engineering guidance, actions, constraints, and
ten-trial budget. All variants select their actions autonomously.

\textbf{Frozen Experience.} This variant uses the current recipe and latest feedback with fixed
guidance, without earlier trial history or online experience updates.
Its maximum EPE is 8.082 nm on Poly and 16.121 nm on Metal1, versus 5.361
and 15.692 nm for Full. Accumulating trial experience improves the final
recipes on both layers, with a larger maximum-EPE improvement on Poly.

\textbf{Domain Response Reflection.} Replacing this component with generic reflection raises
maximum EPE to 6.213 and 16.280 nm. Spatial evidence remains available, but
explicit guidance on interpreting hotspot-level responses and changes in
overall recipe quality is removed.

\textbf{Scoped Experience Revision.} Without this component, factual checks remain but
intervention-topic preservation and evidence-reclassification review are
removed. Maximum EPE rises from 5.361 to 5.475 nm on Poly and from
15.692 to 16.334 nm on Metal1.

\textbf{Experiment Guidance.} Without this component, judgments remain available but explicit strategy
updates and proposed trials are omitted, yielding 5.870 and 16.154 nm.
These comparisons support interpreting measured responses, revising their
applicability, and using them to guide subsequent experiments.
Full has the lowest maximum EPE on both layers and the lowest mean EPE on
Metal1. On Poly, removing revision or guidance lowers mean EPE to 0.632 or
0.624 nm from 0.645 nm, showing that worst-hotspot and average
accuracy need not improve together.

\subsection{Further Analysis}
\label{sec:further-analysis}

\subsubsection{Value of Evolved Experience}
\label{sec:experience-value}

To isolate the value of experience beyond the retained recipe and raw history,
each case resumes from the same fifth-trial recipe,
feedback, and history, using either generic ACE experience curated from those
trials or the recorded VPEvolve experience. Both banks remain fixed through
five further trials with the same model, guidance, and R0 constraints.
VPEvolve experience yields mean additional maximum-EPE reductions of 0.681 nm
on Poly and 2.494 nm on Metal1, versus 0.453 and 2.104 nm for ACE
(Figure~\ref{fig:experience-value}a--b). Final mean EPE is nearly unchanged on
Poly and lower on Metal1 (0.693 versus 0.698 nm). With recipe and history held
fixed, the larger average gains on both layers show that organizing process
feedback into engineering judgments can improve subsequent decisions.
Appendix~\ref{app:continuation-results} reports paired differences and intervals.

\subsubsection{Local Correction after Global Exploration}
\label{sec:local-rule-case}

\begin{figure}[t]
\centering
\includegraphics[width=\textwidth]{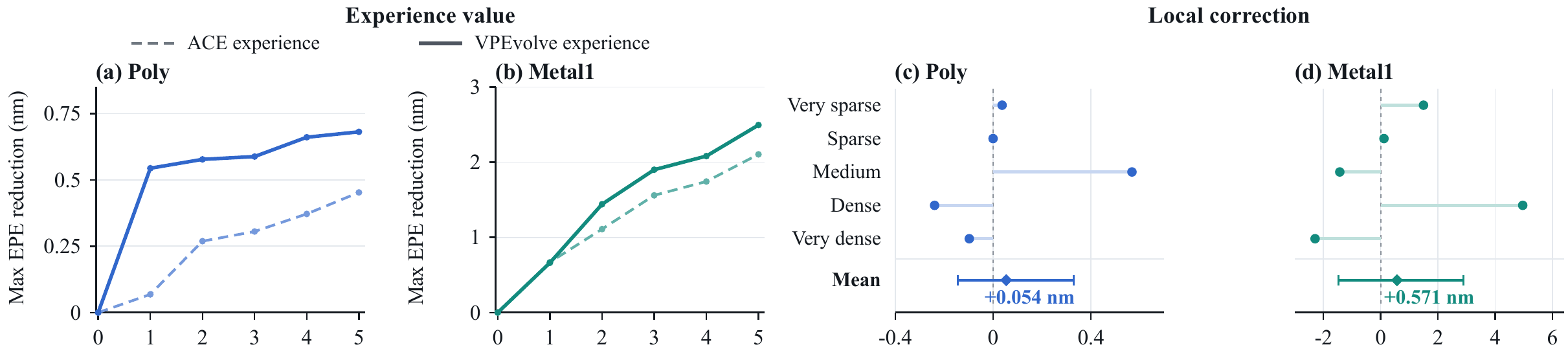}
\caption{\textbf{Value of evolved experience and local correction.}
(a--b) Mean additional maximum-EPE reduction using fixed VPEvolve (solid) or
ACE (dashed) experience from a shared fifth-trial recipe and history.
(c--d) Paired maximum-EPE advantage after four trials; positive values favor
local actions. Points are cases, diamonds are means, and bars are 95\%
paired bootstrap intervals.}
\label{fig:experience-value}
\label{fig:local-correction-paired}
\end{figure}

To test whether spatial actions complement global tuning, we select one case
per density stratum and layer. Paired continuations start immediately before
the first local intervention, sharing recipe, feedback, and experience.
Both update experience over four trials; one permits global edits alone,
the other also permits local correction. Allowing local actions lowers mean
maximum EPE from 18.583 to 18.012 nm on Metal1 and from 5.238 to 5.185 nm on
Poly (Figure~\ref{fig:local-correction-paired}c--d). It improves three of five
Metal1 cases and two of five Poly cases, with one Poly tie. Mean EPE is higher
with local actions (0.651 versus 0.635 nm on Poly; 0.698 versus 0.630 nm on
Metal1), as is PVB, while both branches remain feasible. Local corrections
provide another way to reduce residual worst errors after global exploration,
with their value depending on the remaining hotspots and available global edits.

\begin{figure}[t]
\centering
\includegraphics[width=\textwidth]{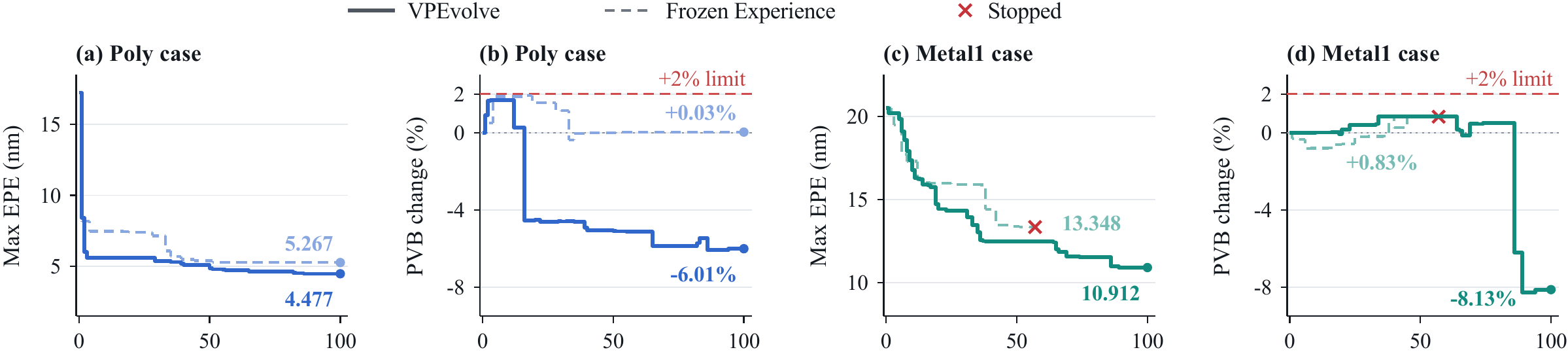}
\caption{\textbf{Long-horizon recipe evolution.}
Retained maximum EPE (a, c) and PVB change from R0 (b, d) for Poly (blue)
and Metal1 (green). VPEvolve is solid; Frozen Experience is dashed.
Red dashes mark the +2\% PVB limit; crosses mark Frozen's termination
at trial 57 on Metal1. Evaluated candidates count
toward the horizon even when not retained. Endpoint repeats are excluded.}
\label{fig:long-horizon-comparison}
\end{figure}

\subsubsection{Long-Horizon Optimization}
\label{sec:long-horizon}

We extend the budget to 100 trials on one case per layer to test whether
experience evolution supports sustained optimization. Full and Frozen
Experience start from R0 with empty banks and shared guidance, using
DeepSeek-V4.1-Flash through its API under the original quality constraints
(Figure~\ref{fig:long-horizon-comparison}). On Poly, both complete 100 trials;
Full reaches 4.477 nm maximum EPE versus 5.267 nm for Frozen, with mean EPE
0.639 versus 0.813 nm. PVB changes by $-6.01\%$ and $+0.03\%$, respectively.

On Metal1, Frozen repeatedly requests information and terminates after 57
trials. At that shared horizon, Full already has lower maximum EPE
(12.470 versus 13.348 nm). It continues to 100 trials, reaching 10.912 nm,
with mean EPE 0.529 nm and PVB 8.13\% below R0. The comparison reveals both
better quality at the shared horizon and further gains as Full continues.
Together, these trajectories support evolving experience as a means of
sustaining useful recipe refinement over longer interactions.
Appendix~\ref{app:long-search} gives completion and recovery details.

\vspace{-6pt}
\subsubsection{Generalization across Language Models}
\label{sec:final-cross-model}

\begin{figure}[t]
\centering
\includegraphics[width=\textwidth]{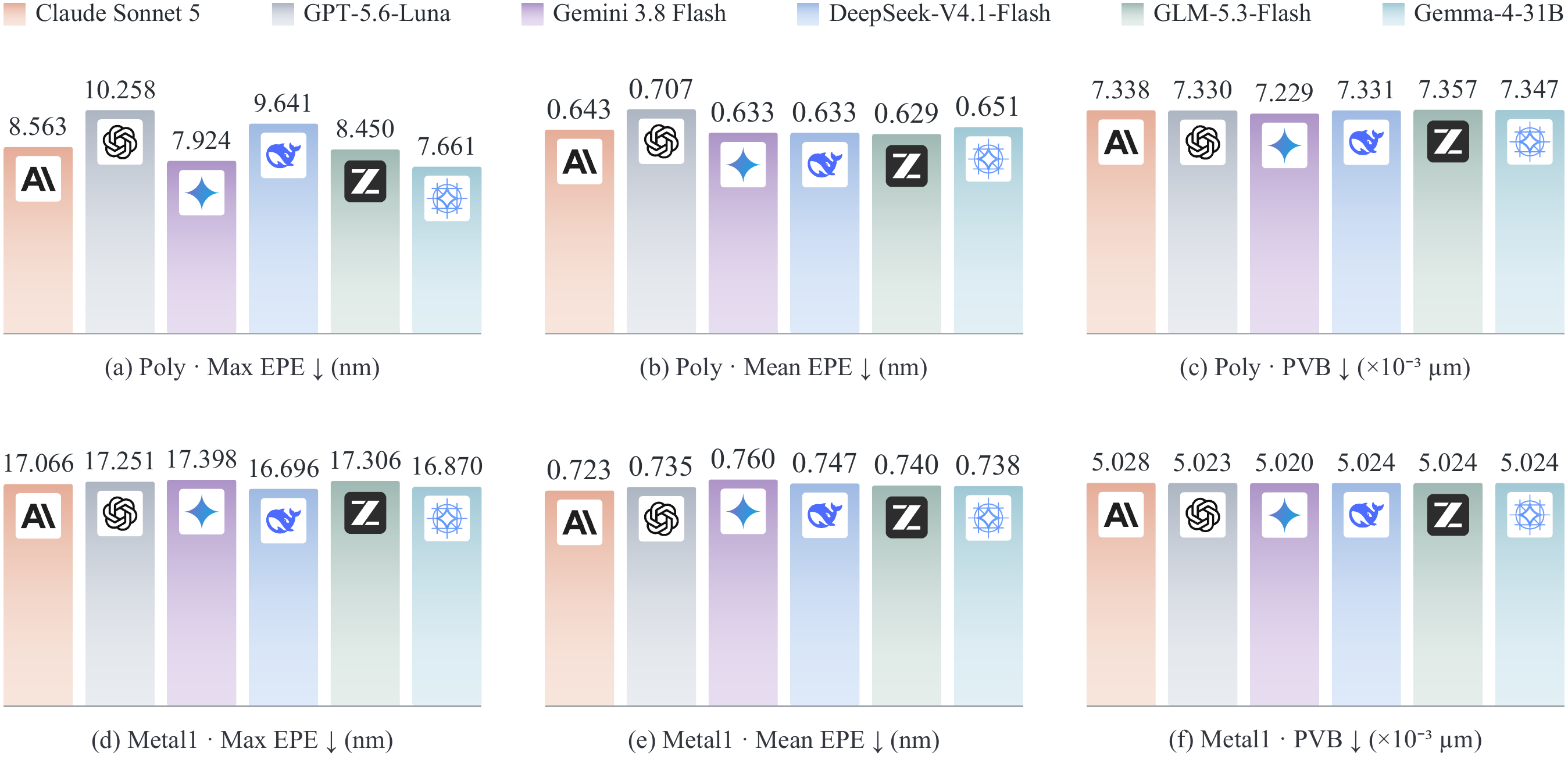}
\caption{\textbf{VPEvolve across language models.}
Each model serves as actor, reflector, and curator through its API.
Bars show benchmark means for Poly (top) and Metal1 (bottom).
EPE is in nm and PVB in $10^{-3}\,\mu$m; lower is better.}
\label{fig:cross-model-actor}
\vspace{-10pt}
\end{figure}

To test whether VPEvolve depends on a particular language model, we replace
all three roles together with each of six models: Claude, GPT, Gemini,
DeepSeek, GLM, and Gemma, accessed through their APIs
(Figure~\ref{fig:cross-model-actor}). All six lower mean and maximum EPE from
R0. Mean maximum EPE ranges from 7.661 to 10.258 nm on Poly and 16.696 to
17.398 nm on Metal1, versus initial values of 18.294 and 22.052 nm.
Gemma and DeepSeek achieve the lowest maximum EPE on Poly and Metal1;
GLM and Claude lead on mean EPE. PVB ranges from 7.229 to 7.357 on Poly
and 5.020 to 5.028 on Metal1, in $10^{-3}\,\mu$m. The recipe and experience loop improves quality across model families,
with rankings varying by layer and metric.
Appendix~\ref{app:cross-model-interface} gives the inference settings.

\vspace{-6pt}
\section{Conclusion}
\label{sec:discussion}

Each lithography trial can improve a recipe and leave evidence for the next
engineering decision. \system makes both outputs part of the same
development loop. Its Virtual Process Engineer harness gives a frozen LLM
process guidance, layout analysis, recipe editing, and commercial-tool
evaluation; its Skill Bank uses LLM reflection and curation to turn trial
feedback into scoped, evidence-linked judgments. The commercial-tool
benchmark evaluates the complete workflow,
and the released FreePDK45 layouts and simulator support open simulation
studies. We hope this work helps process engineers build on the experience
earned through repeated trials and makes the demanding work of recipe
development a little easier.

\pagebreak
\FloatBarrier

\bibliography{references}
\bibliographystyle{template/iclr2027_conference}

\clearpage
\appendix
\FloatBarrier
\setcounter{table}{0}
\renewcommand{\thetable}{A\arabic{table}}
\renewcommand{\theHtable}{A\arabic{table}}
\setcounter{figure}{0}
\renewcommand{\thefigure}{A\arabic{figure}}
\renewcommand{\theHfigure}{A\arabic{figure}}
\setcounter{topnumber}{3}
\section{Layouts and Dataset Construction}
\label{app:dataset}

The benchmark uses windows cropped from full-chip designs generated with the
FreePDK45 technology library \citep{freepdk45}, covering Poly (layer 9/0)
and Metal1 (layer 11/0). Each layer contains ten windows, with two from each
of five density strata ranging from very sparse to very dense. Each window
serves as one evaluation unit in the main comparison.

Every window has a fixed $10\times10\,\mu$m scoring core and a $2\,\mu$m
context halo. Whole polygons intersecting the context are retained in their
original coordinates to supply simulation context; only the core is scored.
The sampling manifest records the source-layout hash, layer, density group,
and core and context coordinates. The local-correction study uses one case
per stratum in each layer. The benchmark layouts and their metadata are
included in the public release.

\section{Experimental Platform and Evaluation Protocol}
\label{app:platform}

\paragraph{Evaluation platform.}
\label{app:operational-boundary}
Experiments use licensed Calibre software with two threads per job and fix
the process model, rule decks, scoring code, available edits, and evaluation
budget. Candidate validation follows Section~\ref{sec:method}: the harness
checks the exact parent recipe, command syntax, and allowed edits, with
pattern-selection and spatial-motion checks for local corrections. The
actor selects trials, and measured responses guide recipe retention and
experience updates. The public release includes a standalone lithography
simulator for independent layout-level studies.

\paragraph{Measurements and selection.}
\label{app:metrics}
Normal edge-placement error (EPE) is measured at fixed external gauges with
1-nm spacing, excluding the first and last 25 nm of each original edge.
Measurements use the $10\times10\,\mu$m scoring core. Max and Mean EPE use
absolute errors at nominal focus and dose; Top-10 EPE averages the ten
spatially separated worst hotspots. PVB is process-variation-band area per
target-edge length. A feasible recipe has zero MRC and SRAF-printing
violations and keeps mean EPE and PVB within 102\% of their R0 values.
Selection minimizes maximum EPE, breaking ties by Top-10 EPE, Mean EPE,
and PVB, in that order.

For the main and mechanism comparisons, success requires completion with a
feasible maximum-EPE reduction of at least 0.1 nm relative to R0. Success
rates count all assigned runs. Layer-wise and overall quality means weight
assigned windows equally, using each run's last retained feasible recipe;
missing measurements are not imputed.

\paragraph{Evaluation budgets.}
Table~\ref{tab:appendix-budgets} lists the trial budgets. Global, local, and
diagnostic trials share the same budget, including evaluated candidates that
are not retained and attempts with execution or measurement failures.
Proposals rejected before evaluation consume no OPC evaluation. R0
initialization and an endpoint repeat are separate from search; the repeat
re-evaluates the retained recipe to check its metrics and output geometry.
Appendix~\ref{app:implementation} specifies model settings, and
Appendix~\ref{app:runtime} reports evaluation and model-call costs.

\begin{table}[t]
\caption{Experimental configurations and per-run trial budgets. Each
configuration uses all listed windows; continuations count additional trials.
Budgets exclude initialization and one final re-evaluation per run
(per completed run for long-horizon evaluation).}
\label{tab:appendix-budgets}
\centering\small
\begin{tabular}{lccc}
\toprule
Experiment & Windows & Configurations & Trial budget \\
\midrule
Main comparison & 20 & 8 methods & 10 \\
Mechanism ablation & 20 & 5 variants & 10 \\
Cross-model evaluation & 20 & 6 models & 10 \\
Experience-value continuation & 20 & 2 experience banks & 5 \\
Local-action continuation & 10 & 2 action sets & 4 \\
Long-horizon evaluation & 2 & 2 variants & 100 \\
\bottomrule
\end{tabular}
\end{table}

\label{app:reproducibility}
Trial records link compiled candidates, parent recipes, edits, measurements,
feasibility, and retention decisions. Run metadata records window coordinates
and hashes, initial recipe and experience, model settings, prompts, tool
manifests, code revisions, evaluator hashes, and call counts.

\section{Baseline Details}
\label{app:baselines}

The baselines compare direct action selection, interaction history,
accumulated experience, and numerical search under shared initial recipes,
evaluation budgets, and quality constraints. Agent baselines use the same
recipe-editing actions.

\noindent\textbf{Raw Actor.}
Proposes recipe edits from the current task state and measurements, using
the shared tools and evaluator without consolidating trial outcomes into
an explicit experience bank.

\noindent\textbf{ReAct} \citep{yao2023react}.
Interleaves reasoning, actions, and observations. During recipe optimization,
it uses interaction history and tool feedback to revise its working
hypothesis and choose subsequent edits, without a separate experience bank.

\noindent\textbf{Textual Memory} \citep{shinn2023reflexion,zhao2024expel}.
Accumulates natural-language reflections on previous attempts. Trial outcomes
and failures are summarized chronologically, providing textual experience
that guides the agent's subsequent recipe edits.

\noindent\textbf{ReasoningBank} \citep{ouyang2026reasoningbank}.
Distills previous attempts into reusable lessons with applicability conditions
and potential pitfalls. Our adaptation retrieves relevant lessons using
lexical overlap with the current case and hotspot feedback to guide recipe edits.

\noindent\textbf{ACE} \citep{zhang2026ace}.
Maintains an evolving playbook through reflection and incremental curation.
The reflector analyzes recipe-trial feedback, and the curator incorporates
these observations into playbook entries that guide subsequent actions.

\noindent\textbf{Static Harness} \citep{pi2026agent}.
Uses a Pi-style agent harness with fixed engineering guidance. The actor
responds to new measurements when choosing recipe edits, while the process
guidance remains unchanged throughout optimization.

\noindent\textbf{Bayesian Optimization} \citep{snoek2012practical,gardner2014constrainedbo}.
Searches model-based recipe parameters using Gaussian-process surrogates
for quality and feasibility. Constrained expected improvement selects the
next candidate, and each evaluation updates the surrogates.

\section{Implementation and Experimental Settings}
\label{app:implementation}

\paragraph{Models and initialization.}
Unless otherwise specified, all LLM roles use Qwen3.6-27B served locally on
four NVIDIA H100 GPUs, with a 131,072-token context, 4,096 output tokens,
and temperature zero. These settings apply to the main comparison,
mechanism ablations, and paired continuations. VPEvolve and its mechanism
ablations start from R0 with an empty Skill Bank and shared engineering
guidance, under the budgets in Table~\ref{tab:appendix-budgets}.

\begin{algorithm}[t]
\caption{VPEvolve: Recipe Optimization with Evolving Experience}
\label{alg:loop}
\small
\definecolor{algRecipe}{HTML}{285E95}
\definecolor{algExperience}{HTML}{197D74}
\definecolor{algSelection}{HTML}{566273}
\renewcommand{\algorithmicrequire}{\textbf{Input:}}
\renewcommand{\algorithmicensure}{\textbf{Output:}}
\begin{algorithmic}[1]
\REQUIRE Layout $x$, initial recipe $r_0$ and feedback $e_0$, trial budget $B$;\\
frozen model $\pi_\theta$, VPE harness $H$, lithography evaluator $F$
\ENSURE Retained recipe $r_B$, Skill Bank $\mathcal E_B$, trial history $\mathcal T_B$
\STATE Initialize $\mathcal E_0\gets\varnothing$, $\mathcal T_0\gets\varnothing$
\FOR{$t=0,\ldots,B-1$}
  \STATE \textcolor{algRecipe}{$\triangleright$ \textit{Actor: choose a global, local, or diagnostic trial}}
  \STATE $V_t\gets\operatorname{Retrieve}(\mathcal E_t,x,r_t,e_t)$
  \STATE $a_t\sim\pi^{\mathrm{act}}_{\theta,H}(\cdot\mid x,r_t,e_t,V_t,\mathcal T_t)$
  \STATE Validate and compile $\widetilde r_t\gets\operatorname{Apply}(r_t,a_t)$
  \STATE Evaluate $\widetilde e_t\gets F(x,\widetilde r_t)$; record $\tau_t=(r_t,a_t,\widetilde r_t,e_t,\widetilde e_t)$
  \STATE $\mathcal T_{t+1}\gets\mathcal T_t\cup\{\tau_t\}$
  \STATE \textcolor{algSelection}{$\triangleright$ \textit{Recipe update: retain only feasible improvements}}
  \STATE $(r_{t+1},e_{t+1})\gets\operatorname{Select}(r_t,e_t,\widetilde r_t,\widetilde e_t)$
  \STATE \textcolor{algExperience}{$\triangleright$ \textit{Reflector: Domain Response Reflection}}
  \STATE $h_t\gets\operatorname{Reflect}_{\pi_\theta}(\tau_t)$
  \STATE \textcolor{algExperience}{$\triangleright$ \textit{Curator: Scoped Experience Revision and Experiment Guidance}}
  \STATE $\Delta_t\gets\operatorname{Curate}_{\pi_\theta}(\mathcal E_t,h_t,\tau_t)$
  \STATE $\mathcal E_{t+1}\gets\operatorname{CheckApply}(\mathcal E_t,\Delta_t,\tau_t)$
\ENDFOR
\STATE Re-evaluate $r_B$ to verify the retained endpoint
\end{algorithmic}
\end{algorithm}

\label{app:cross-model-interface}
In the cross-model study, all three roles use the same evaluated model
through its API: Claude Sonnet 5 \citep{anthropic2026sonnet5},
GPT-5.6-Luna \citep{openai2026gpt56},
Gemini 3.8 Flash \citep{google2026gemini38flash},
DeepSeek-V4.1-Flash \citep{deepseek2026v41flash},
GLM-5.3-Flash \citep{zai2026glm53flash}, or
Gemma-4-31B \citep{gemmateam2026gemma4}.
All use a 65,536-token context; Table~\ref{tab:model-settings} lists generation
settings. Reasoning is disabled for Claude and GPT and set to low for Gemini
and GLM. The sampling seed is fixed where supported. Long-horizon runs use
DeepSeek-V4.1-Flash for all three roles through its API, with temperature zero,
reasoning disabled, a 131,072-token context, and 4,096 output tokens.

\begin{table}[t]
\caption{Generation settings for the cross-model comparison. Output limits
include reasoning tokens. Default denotes the provider's sampling setting.}
\label{tab:model-settings}
\centering\small
\begin{tabular}{lrr}
\toprule
Model & Output tokens & Temperature\\
\midrule
Claude Sonnet 5 & 4,096 & Default\\
GPT-5.6-Luna & 4,096 & Default\\
Gemini 3.8 Flash & 4,096 & 0.3\\
DeepSeek-V4.1-Flash & 4,096 & 0.3\\
GLM-5.3-Flash & 16,384 & 0.3\\
Gemma-4-31B & 4,096 & 0.3\\
\bottomrule
\end{tabular}
\end{table}

\paragraph{Experience updates and controls.}
\label{app:experience-review}
Algorithm~\ref{alg:loop} summarizes the optimization loop. Reflection compares
the edit and predicted response with before/after measurements at fixed
hotspots, new worst locations, and aggregate quality metrics, distinguishing
an ineffective correction from a repaired target followed by a new limiting
hotspot. Curation adds, updates, splits, or merges judgments linking
interventions and applicability to supporting trials and counterexamples.
Revisions preserve the intervention topic, review evidence reclassification
or removal, and check cited interventions against trial records. Each
judgment may propose a strategy update and up to two experiments with
predicted effects and observations that would falsify the explanation.

Frozen Experience fixes the initial bank and guidance and uses only the
current recipe and latest feedback. The three ablations remove domain
reflection guidance, topic-preservation and evidence-reclassification review,
or strategy updates and proposed experiments. All experience-update
conditions share actor guidance, available actions, physical selection
criteria, and factual citation checks.

\section{Paired Continuation Studies}
\label{app:local-rule}
\label{app:continuation-results}

The paired continuations in Sections~\ref{sec:experience-value}
and~\ref{sec:local-rule-case} hold the starting recipe and history fixed to
compare experience quality and the value of local actions. Both studies use
the default model settings in Appendix~\ref{app:implementation}, the R0
quality constraints, and an endpoint repeat to verify the retained recipe.

\paragraph{Experience value.}
All 20 cases resume from Full's fifth-trial recipe, feedback, and history.
ACE curates the five preceding observations from the same initial knowledge;
VPEvolve supplies its recorded experience. Both banks remain fixed during
five further trials, sharing the model, engineering guidance, and access to trial history.

\paragraph{Local actions.}
One case per density stratum and layer is selected before observing outcomes,
giving five Poly and five Metal1 cases. Both branches start immediately before
the first local intervention with the same recipe, feedback, and experience,
and update experience over four trials. One permits global edits alone; the
other permits global and local edits. Local actions displace selected mask
fragments at a diagnosed edge while keeping target geometry and evaluation
gauges fixed. The selected region and displacement are checked before
execution under the shared quality constraints.

\begin{table}[t]
\caption{Endpoint quality in the paired continuation studies. EPE is in nm; PVB is in $10^{-3}\,\mu$m. Experience conditions cover the benchmark; each local-action condition has five cases per layer.}
\label{tab:continuation-quality}
\centering\small
\setlength{\tabcolsep}{4pt}
\begin{tabular}{lrrrrrr}
\toprule
& \multicolumn{3}{c}{Poly} & \multicolumn{3}{c}{Metal1}\\
\cmidrule(lr){2-4}\cmidrule(l){5-7}
Condition & Max$\downarrow$ & Mean$\downarrow$ & PVB$\downarrow$ & Max$\downarrow$ & Mean$\downarrow$ & PVB$\downarrow$\\
\midrule
ACE experience & 5.313 & 0.647 & 7.404 & 16.328 & 0.698 & 5.022\\
VPEvolve experience & 5.085 & 0.646 & 7.398 & 15.938 & 0.693 & 5.021\\
\midrule
Global only & 5.238 & 0.635 & 7.375 & 18.583 & 0.630 & 4.980\\
Global + local & 5.185 & 0.651 & 7.379 & 18.012 & 0.698 & 4.996\\
\bottomrule
\end{tabular}
\end{table}

\begin{table}[t]
\caption{Paired maximum-EPE advantage of VPEvolve experience over ACE experience, or of global and local actions over global actions alone. Positive values favor the first condition. Intervals use 20,000 bootstrap resamples of paired cases.}
\label{tab:continuation-paired}
\centering\small
\begin{tabular}{llrrr}
\toprule
Comparison & Layer & Pairs & Mean (nm) & 95\% interval (nm)\\
\midrule
Experience curation & Poly & 10 & 0.228 & [-0.098, 0.735]\\
Experience curation & Metal1 & 10 & 0.390 & [-0.207, 1.022]\\
Local actions & Poly & 5 & 0.054 & [-0.143, 0.329]\\
Local actions & Metal1 & 5 & 0.571 & [-1.465, 2.883]\\
\bottomrule
\end{tabular}
\end{table}

Table~\ref{tab:continuation-quality} reports endpoint quality;
Table~\ref{tab:continuation-paired} gives paired maximum-EPE differences with
95\% intervals from 20,000 bootstrap resamples of paired cases.
VPEvolve experience and local actions have positive mean advantages on both
layers. Local actions also increase mean EPE and PVB, reflecting a tradeoff
within the feasibility limits.

\begin{table}[t]
\caption{Mean calls per assigned window in the main comparison, obtained by
dividing Table~\ref{tab:current-online}'s totals by 20. Commercial-tool calls
include endpoint repeats and exclude R0 initialization.}
\label{tab:call-costs}
\centering\small
\begin{tabular}{lrr}
\toprule
Method & Commercial-tool calls & Model requests\\
\midrule
Raw Actor & 11 & 22.75\\
ReAct & 11 & 26.20\\
Textual Memory & 11 & 51.50\\
ReasoningBank & 11 & 49.40\\
ACE & 11 & 91.65\\
Static Harness & 11 & 25.10\\
Bayesian optimization & 11 & 0.00\\
\midrule
VPEvolve & 11 & 64.75\\
\bottomrule
\end{tabular}
\end{table}

\section{Evaluation and Model-Call Cost}
\label{app:runtime}

Table~\ref{tab:call-costs} expresses the main comparison's call counts per
assigned window. All methods use 220 commercial-tool evaluations across
20 windows: ten optimization trials and one endpoint repeat per window.
These counts exclude R0 initialization and include failed evaluations
(Appendix~\ref{app:platform}).

Model requests cover action selection, tool use, repairs, reflection, and
curation. Per window, VPEvolve uses 64.75 requests, compared with 26.20 for
ReAct and 91.65 for ACE.

\section{Long-Horizon Evaluation}
\label{app:long-search}
\label{app:long-horizon-execution}

The 100-trial comparison (Section~\ref{sec:long-horizon}) starts both variants
from R0 with empty Skill Banks and shared guidance on one case per layer.
It uses the DeepSeek-V4.1-Flash API settings in
Appendix~\ref{app:implementation} and the quality constraints and evaluation
accounting in Appendix~\ref{app:platform}.

Both Poly runs and Metal1 Full complete 100 trials. On Metal1, Frozen
repeatedly requests evidence excluded by its history restriction after
trial 56; recovery completes one more trial before termination at trial 57.
Poly Full also repeats evidence requests after trial 97 and completes the
run following recovery. Recovery extends model-call allowances, with actual
costs retained. The Metal1 comparison measures quality at trial 57 and the
further gains as Full continues. Across both layers, Full and Frozen
complete 200 and 157 trials (357 total), excluding initialization and
endpoint repeats.
\FloatBarrier

\end{document}